\pdfoutput=1
\documentclass[11pt]{article}

\usepackage[]{acl}

\usepackage{times}
\usepackage{latexsym}
\usepackage{todonotes}
\usepackage{multirow}

\usepackage[T1]{fontenc}
\usepackage[utf8]{inputenc}

\usepackage{microtype}

\newcommand\blfootnote[1]{%
  \begingroup
  \renewcommand\thefootnote{}\footnote{#1}%
  \addtocounter{footnote}{-1}%
  \endgroup
}
\title{Bias at a Second Glance: A Deep Dive into Bias for German Educational Peer-Review Data Modeling}

\author{Thiemo Wambsganss$^*$\textsuperscript{1}, Vinitra Swamy$^*$\textsuperscript{1}, Roman Rietsche\textsuperscript{2},\\ \textbf{Tanja Käser\textsuperscript{1}} \\
  \textsuperscript{1} EPFL, Lausanne, Switzerland\\
\footnotesize{ {\tt \{thiemo.wambsganss, vinitra.swamy, tanja.kaeser\}{\tt @epfl.ch}}}\\
  \textsuperscript{2}  University of St.Gallen, St.Gallen, Switzerland\\
  \footnotesize{ \tt {roman.rietsche@unisg.ch}}\\
\\}  
  
\begin{document}
\maketitle
\blfootnote{$^\ast$ These authors contributed equally to this work.}
\begin{abstract}
%Targets UN Sustainability objectives: 4 (quality education), 5 (gender equality), 10 (reduced inequalities).
%Natural Language Processing (NLP) and Machine Learning (ML) models have been increasingly utilized for advancing adaptivity of educational applications.
%The use of text data for the educational machine learning tasks has become increasingly popular across feedback systems, interactive tutors, and automated essay grading. However, recent research in the natural langauge processing (NLP) community has highlighted a variety of biases in pre-trained language models. Existing studies investigate bias in different domains but are limited when it comes to more fine-grained analysis on educational corpora, especially in a language that is not English. 
Natural Language Processing (NLP) has become increasingly utilized to provide adaptivity in educational applications. However, recent research has highlighted a variety of biases in pre-trained language models. While existing studies investigate bias in different domains, they are limited in addressing fine-grained analysis on educational and multilingual corpora.
In this work, we analyze bias across text and through multiple architectures on a corpus of 9,165 German peer-reviews collected from university students over five years. Notably, our corpus includes labels such as helpfulness, quality, and critical aspect ratings from the peer-review recipient as well as demographic attributes. We conduct a Word Embedding Association Test (WEAT) analysis on (1) our collected corpus in connection with the clustered labels, (2) the most common pre-trained German language models (T5, BERT, and GPT-2) and GloVe embeddings, and (3) the language models after fine-tuning on our collected data-set. In contrast to our initial expectations, we found that our collected corpus does not reveal many biases in the co-occurrence analysis or in the GloVe embeddings. However, the pre-trained German language models find substantial conceptual, racial, and gender bias and have significant changes in bias across conceptual and racial axes during fine-tuning on the peer-review data. With our research, we aim to contribute to the fourth UN sustainability goal (quality education) with a novel dataset, an understanding of biases in natural language education data, and the potential harms of not counteracting biases in language models for educational tasks.

\end{abstract}

\section{Introduction} 

%motivation
In recent years, Natural Language Processing (NLP) and Machine Learning (ML) have been extensively used for improving adaptivity and individualization of educational technology \cite{Rose2008AnalyzingLearning,Xu2021FromAI}. Researchers and practitioners have been developing a plethora of writing support systems \cite{song2014applying,Lauscher2019ArguminSci:Writing} and conversational agents \cite{Ruan2019QuizBot:Knowledge,Weber2021PedagogicalEducation}. More generally, there has been a rise in intelligent tutoring systems for educational purposes which provide learners adaptive feedback, e.g., on grammatical structures \cite{white-rozovskaya-2020-comparative,katinskaia-yangarber-2021-assessing,kerz-etal-2021-automated}, language learning \cite{putra-etal-2021-parsing}, argumentation \cite{song2014applying,Lauscher2019AnPublications}, or even empathy skills \cite{Wambsganss2021SupportingStudents}.
%problem

The technology for language-based personalization in education comes with a cost; a large body of research has been investigating and revealing biases in NLP systems \cite{bolukbasi2016man,Sun.2019}. Bias has been found in multiple steps along the general NLP pipeline including the task setting, training data, pre-trained models (e.g. word embeddings), and fine-tuned algorithms \cite{Schramowski.382021,Sun.2019,caliskan2017semantics,bolukbasi2016man}, shedding a darker light on the simple usage of these models for human-centered applications, especially in education. NLP systems containing bias in any of these parts of the modeling pipeline can produce gender, racially, or conceptually biased predictions and amplify biases present in the underlying training sets (e.g., \newcite{Baker2021AlgorithmicEducation,hutchinson2019,Sun.2019}). The propagation of gender bias in NLP algorithms poses the danger of reinforcing damaging stereotypes in downstream applications, e.g., for automatic essay scoring \cite{Ostling2013AutomatedSwedish,Yannakoudakis2011ATexts}.
%Especially for a digitized educational system with automatic essay scoring at high schools \cite{Ostling2013AutomatedSwedish,Yannakoudakis2011ATexts,Blanchard2013Toefl11:English} or writing assistants in Massive Open Online Course (MOOCs) \cite{Seaman2018HigherGroup}, this can have significant impact on the quality and equality of our societies.
\begin{figure*}[!htb]
\centering
 \includegraphics[width=\textwidth]{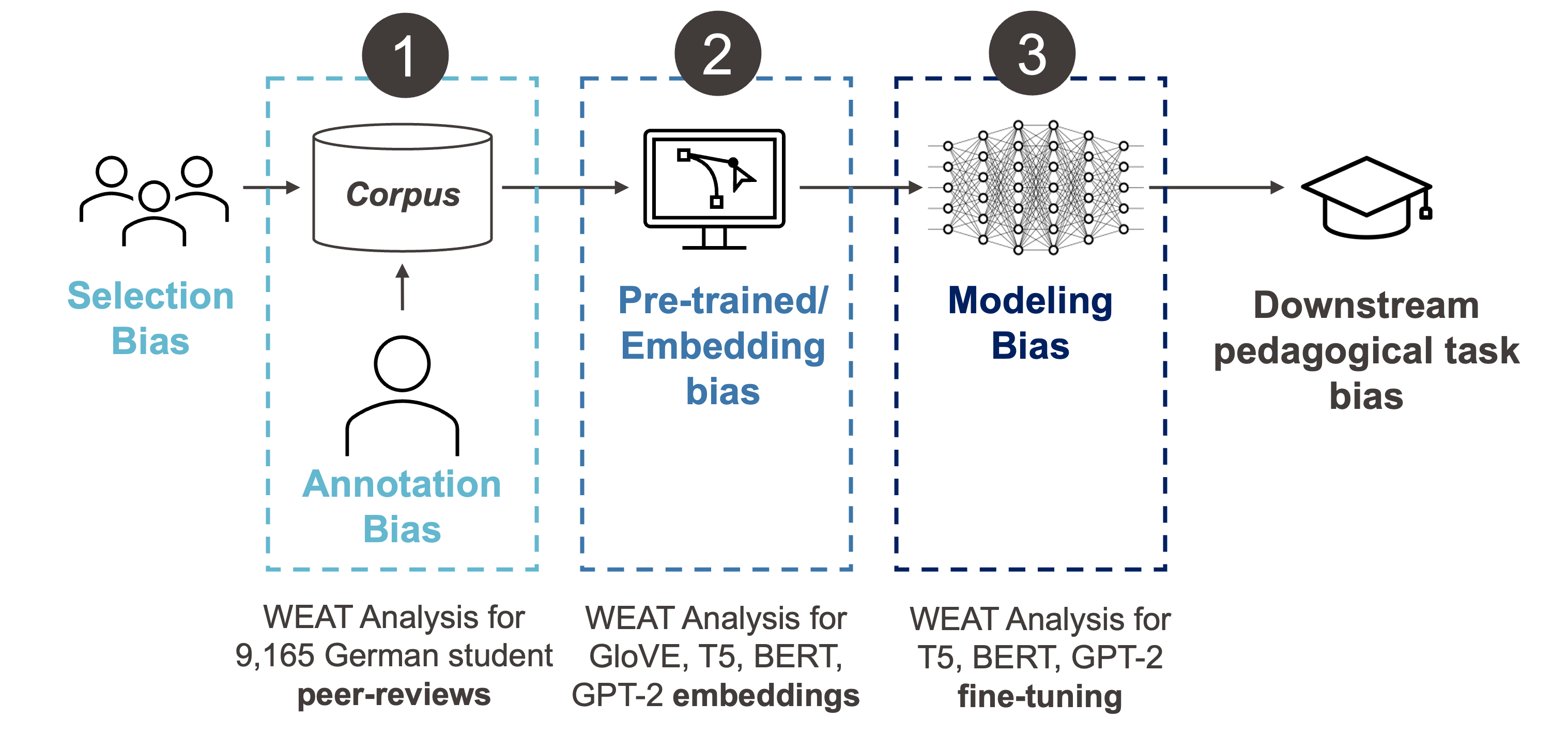}
 \caption{Overview of evaluating biases in educational natural language data along the NLP pipeline for pedagogical downstream tasks following \newcite{Hovy2021}. We analyzed a data set of 9,165 German peer-reviews in combination with the most common pre-trained language models (T5, GPT-2, BERT) and GloVe embeddings before and after fine-tuning with the WEAT analysis for conceptual, racial, and gender biases. 
 \label{fig:concept}
}
\end{figure*}

%related work and gap --> we need more!
While prior research on bias in education has mostly focused on non-language based interaction data, several recent reviews have called for extending the investigations of fine-grained bias analysis on educational corpora (e.g., \newcite{Baker2021AlgorithmicEducation,Blodgett2020}). Recent work, for example, has shown negative impact of gender bias on CV screening \cite{Andersson2021} or of algorithmic racial bias in child welfare programs \cite{cheng2022child}. There are only few works looking at detailed bias in educational natural language data outside of English language corpora and North American context \cite{Baker2021AlgorithmicEducation}. %\cite{Sun.2019} have looked in how to mitigate gender bias in general NLP tasks. \cite{schramowski2022large} investigated moral judgments of large language models. However, deep analysis of educational texts, as well as, on pre-trained and fine-tuned models in languages other than English is still scarce. 
For instance, \newcite{Baker2021AlgorithmicEducation} states the need to investigate "the differences in the performance of essay scoring algorithms for different racial groups". However, as they found, "this possibility has not yet been systematically investigated in the published literature" \cite{Baker2021AlgorithmicEducation}.
%Objective
Hence, our objective is to address this gap in research and to take a deep dive into one exemplary pedagogical scenario which includes heavy language data: student peer-reviews. Student peer-reviewing is a modern domain-independent pedagogical scenario which has been increasingly used to annotate corpora, analysis of feedback texts with trained models, and provide students feedback with adaptive applications \cite{Nicol.2014}, e.g., for argumentation skill training \cite{Wambsganss2020ALSkills} or empathy skills \cite{Wambsganss2021SupportingStudents,Wambsgan2022ELEASkills}.

%method
In order to conduct a rigorous bias analysis, we collected a novel corpus of 9,165 German student peer-reviews of business model feedback. We relied on Word Embedding Association Test (WEAT) analysis \cite{caliskan2017semantics} and the German adaptation of WEAT \cite{kurpicz2020cultural} as a commonly used methodology to assess conceptual, racial and gender bias in different parts of the NLP pipeline \cite{Hovy2021}. Our methodology for analysing the bias is three-fold (see Figure \ref{fig:concept}): (1) we analyse the collected corpus for different bias dimensions to find out if the student-writings already come with bias towards the perceived helpfulness of a review, (2) we assess the most common German language models (T5, BERT and GPT-2) as well GloVe embeddings before fine-tuning them on our data, (3) we fine-tune T5, BERT and GPT-2 on our collected data-set and repeat the WEAT analysis to investigate how the representations have been changed. 

Contrary to our expectations, we found that our collected corpus does not reveal many biases in using WEAT co-occurrence analysis or GloVe models; however, the pre-trained German language models not only come with substantial conceptual, racial, and gender bias but also seem to increase the bias when fine-tuning on our corpus. Our results suggest to (1) do more fine-grained analyses of bias for subsets of data that are significant, (2) examine the bias in pre-trained models before using them, and (3) investigate multilingual data bias more precisely.
Hence, we contribute to literature on bias of educational language data by providing a detailed analysis of one particular but increasingly used pedagogical scenario (peer-reviewing). We contribute our collected corpus of peer-reviews in German for further analysis and hope to provide researchers and practitioners with a detailed analysis and discussion of bias in NLP for education. Finally, we aim to contribute to the UN sustainability goal four for a high quality education and fair (digital) education for all.

%%% Link why peer-reviewing is important
%We show educational biases based on the example of peer-reviewing. Peer-reviews are a modern learning scenario in mostly used in large-scale lectures, enabling students to reflect on their content, receive individual feedback from peers and thus deepen their understanding of the content \cite{Nicol.2014, rietsche2019insights}. It can easily be set up in traditional large-scale learning scenarios or in the growing field of distance-learning scenarios such as massive open online courses (MOOCs), in which often thousands of students are enrolled \cite{Shah.2021, seaman2018grade}.

%For example \newciteauthor{Adamopoulos.2013} \newciteyearpar{Adamopoulos.2013}showed that using peer-reviews within MOOCs have a significant positive effect on course retention of students and a partial effect on the probability that a student successfully completes the course. Additionally, peer-reviewing itself has the challenge that the students who are providing the feedback are non-experts and in most cases, do not experience training on avoiding biases \cite{Double.2020}. In contrast, lecturers usually receive dedicated training in their professional education to avoid biases i.e. in grading, managing a classroom, or providing feedback to students. Therefore, we propose peer-reviewing as highly relevant setting to study biases in education. 

\section{Theoretical Background}

%In this section, we will discuss the background educational literature and motivation for our work in more detail.

\subsection{Text Bias in Education}

Since the 1960s, the problem of bias in educational applications has been noted, and many parts of today's literature on algorithmic bias and fairness have been anticipated (see review and discussion in \newcite{hutchinson2019,Baker2021AlgorithmicEducation}). In order to investigate bias, it is important to define what perspective on bias we take, as many definitions exist in the literature. In our research, we "focus on studying algorithmic bias in terms of situations where model performance is substantially better or worse across mutually exclusive groups" \cite[p.~4]{Baker2021AlgorithmicEducation}. We aim to analyse annotation, embedding, and modeling bias throughout the NLP pipeline for educational downstream tasks (see Figure \ref{fig:concept}). 

Most literature has focused on numerical (non-text) data to analyse bias in educational applications. The literature on bias in education has been mostly investigating differences between race, nationality (students’ current national locations), and gender \cite{Baker2021AlgorithmicEducation}.
For example, \newcite{lee2020evaluation} analysed the differences of an unmodified model to an equity-corrected model for predicting course grade of students. They found that the unmodified models perform worse for underrepresented racial and ethnic groups than for White and Asian students. \newcite{Anderson2019AssessingTF} used five different algorithms to discriminate performance between male and female students in a model that predicted six-year college completion. They discovered that male students had greater false negative rates in the algorithms.

Educational corpora play a minor role in investigating bias in NLP systems; the scope of relevant work is limited. The existing research is mostly centered around English language data, missing insights for other languages and cultures \cite{Baker2021AlgorithmicEducation,Sun.2019}. One relevant study from \newcite{loukina-etal-2019-many} investigates bias in automated essay grading over essays authored by individuals from six different nations. Their results indicate that training nation-specific models leads to various skews between groups, increasing algorithmic bias when compared to training on all groups together. In a more general scope, research in computational linguistics and NLP have increasingly investigated bias in natural language processing systems, including work on bias in embedding spaces (e.g., \newcite{caliskan2017semantics,bolukbasi2016man}) language modeling \cite{Lu2020GenderBI}, co-reference resolution \cite{rudinger-etal-2017-social}, machine translation \cite{stanovsky-etal-2019-evaluating}, or sentiment analysis \cite{kiritchenko-mohammad-2018-examining}. %\cite{Sun.2019} looked in how to mitigate gender bias in general NLP tasks. \cite{Schramowski2022LargeDo} investigated moral judgments of large language models and found that large pre-trained language models contain human-like bias of what is right and wrong. 

The analysis of data collected from commonly used pedagogical scenarios in combination with pre-trained language models (especially outside North America) is scarce (e.g., \newcite{Baker2021AlgorithmicEducation,Sun.2019}). We focus on algorithmic bias in the representations of a novel collected data set, in the most common German pre-trained models and German GloVe embeddings, and finally in the fine-tuned models. Our objective goes beyond investigating the effect of NLP bias on educational designs; we aim to contribute to a vision where downstream educational models are unbiased for equitable education (UN Sustainability objective).

\subsection{NLP Research on Peer Reviewing}
To conduct a rigorous and representative bias analysis of educational data from the field, our objective is a domain-independent pedagogical setting. In this vein, we aim to focus on student peer-reviews, since it is a increasingly growing, modern and digitized educational scenario, which has not only been used to foster factual and conceptual learning goals but also more complex skills such as argumentation \cite{Wambsganss2020ALSkills} or empathy \cite{Wambsganss2021SupportingStudents}.

Peer-reviews are defined as the process when students evaluate and make judgments about submissions of their peers and construct written feedback \cite{Nicol.2014}. The structure of this process builds on the principles of the standard peer-reviewing used in academic journals \cite{Ziman.1974}, where each of the papers is assigned to three anonymous reviewers. These reviewers evaluate the quality of the submission and provide feedback about the strengths and weaknesses of the paper and specify ways to improve it \cite{Meadows.1998}. 

Peer-reviews come with the advantage that students need to take two different perspectives: one of the feedback provider and one of the feedback receiver. The role of feedback provider enables students to practice their critical thinking skills, apply criteria and reflect on their own work. Receiving feedback supports the students to focus more on areas that need improvement and develop a reader´s perspective \cite{Wu.2021}. Utilizing peer-reviews enables students to receive timely feedback in both small-scale and large-scale classes, where feedback from lecturers or instructor assessment is often too late to be implemented. With the pedagogical scenario of edit history, it is even possible to directly apply feedback to improve final version of submission \cite{Higgins.2001}. Additionally, with the rise of educational technologies, the peer-review paradigm is increasingly implemented in common learning management systems (e.g. Canvas, Blackboard) and is used by large MOOC providers. However, the direct use of unfiltered feedback by non-experts poses the concern that students are exposed to biases and inaccuracies \cite{Double.2020}. Therefore, past research has already started to analyze peer-review data with standard NLP techniques. 

%% Analyzing Peer-reviews with NLP 
For example, \newcite{Misiejuk.2021} use NLP to understand students' perceptions of peer-reviews. \newcite{Wu.2020} measure the impact of feedback features such as identification, explanations, and suggestions on the likelihood of that the feedback gets implemented. \newcite{Xiao.2020} trained different models based on RNNs, CNNs, LSTMs, GloVe, and BERT to detecting problem statements in peer assessments. \newcite{Zingle.2019} use CNNs and LSTMs to detect actionable suggestions in peer assessments. 
%%% Examples for Downstream applications
Researchers have also started to develop downstream applications based on model predictions to provide adaptive learning feedback. \newcite{Ramachandran.2017} created a tool for automated assessment of the quality of peer-reviews. \newcite{Bauman.2020} designed a recommender framework which uses a trained model to identify aspects of the review texts that correspond to peer-review helpfulness scores. %The model is used to provide real-time recommendations such as "further suggestions should" be added to improve the review helpfulness. Other work conducted by researchers\footnotemark{\ref{blinded}} developed framework to help students to write more argumentative texts based on a trained model.
Several papers use student peer-review data for annotating arguments for argumentation mining (e.g., \newcite{Wambsganss2020AGerman}) or cognitive and emotional empathy structures for empathy modeling \cite{Wambsganss2021SupportingStudents} to provide students with writing assistance in learning applications.

%Moreover peer-reviews are a prominent research field in NLP to collect and annotated corpora and provide students feedback with (e.g., \cite{wambsganss}). Researchers have successfully trained models to provide students with argumentation feedback based on argumentation mining (e.g., \cite{}), or to promote empathy skill learning based on empathy detection (e.g., ).

%%% Contribution
\textit{Although NLP research exists on and around peer-review data, there are only a handful of investigations on bias along the NLP pipeline \cite{Patchan.2018}. Hence, we propose to investigate which biases occur in education data along the NLP pipeline and in particular in our context in peer-reviews.}

\section{Methodology}

\subsection{Data Collection}
\label{sec:data_collection}

Since there are not many suitable corpora available to analyse bias in student peer-reviews that a) contain a large amount of student-written text in one particular domain (e.g., business model feedback), b) consist of a sufficient size to represent different nuances of characteristics in a balanced fashion and c) come with additional scores such as review helpfulness rated by the receiver of the review or demographics for additional analysis (e.g., gender), we decided to collect our own longitudinal data set.

The peer-reviews of our novel dataset were collected over five years at a university in the German speaking area of Europe.\footnote{The data was collected based on the ethical guidelines of our university.} Overall, we compiled a corpus of 9,165 student-generated peer-reviews in which students provide each other feedback on previously developed business models. The peer-reviewing process was conducted in a double-blind manner; thus the feedback provider and receiver were anonymous. %Thus, we can assume that the peer-reviews do not contain biases, based on the characteristics of the feedback recipient, but purely based on the feedback provider. 
Alongside the text data, we collected subsets of ratings regarding the review helpfulness. This data was collected within the peer-review process; when the authors of the assignment receive the peer-reviews, they performed peer backward assessment \cite{Patchan.2016}. In peer backward assessment, students rate the four items (based on \newcite{Li.2010}): (1) \textit{"The feedback I got from the reviewer was helpful"} (2) \textit{"The feedback I got from the reviewer was high quality"} (3) \textit{"The reviewer was able to identify critical aspects in my submission"} and (4) \textit{"The reviewer was able to provide constructive suggestions on his stated critical aspects"} on a 7-point Likert Scale from totally disagree (1) to totally agree (7), with 4 as a neutral value. Additionally, we captured gender and the year of birth of the review writers. 

\subsection{Data Characteristics}

Our dataset consists of first-year master's students majoring in business innovation. The majority of students have German as their native language. The data was collected from 2015 to 2019 and include 9,165 reviews from 610 unique reviewers and 607 reviewees. We collected demographic data at the beginning of each semester; the student population has an average age of 24.6 years old with a standard deviation of 1.7 years. The average percentage of female students across five years is 37.7\%. Students wrote approximately 9 peer reviews per course with an average length of 220 words.

\subsection{Model Architecture}
\label{sec:model-arch}

We examine four German variations of language model architectures in this paper, chosen for their popularity on downstream tasks: GloVe, BERT, T5, and GPT-2. For GloVe architectures, we train the model from scratch for 100 epochs each, using a vector size of 300, window size of 15, and 8 threads for parallelization \cite{glove}. We obtain all three pre-trained models from HuggingFace \cite{wolf2019huggingface} and fine-tune each model for 10 epochs on a Tesla V100 GPU with batch size 8. For GermanBERT, we fine-tune the model using a standard masked language model training objective with masking rate of 15\% \cite{germanbert}. For German T5, we fine-tune the model using the translation task, translating peer-review text from English to German\footnote{Translations were obtained through the Google Translate API and corrected by a native English speaker.} with max source token length of 128 and global seed 42. The pre-trained multilingual T5 model was fine-tuned on the German MLSum dataset \cite{mt5} before being used for our analysis. German GPT-2 was fine-tuned on the text generation objective with block size 128, and 600 warm-up steps \cite{gpt2}. More details can be found directly in our supplementary code repository.

\subsection{Bias Analysis}
% \label{sec:WEAT-preprocess}
To assess bias along the NLP pipeline suggested by \newcite{Hovy2021}, we rely on the Word Embedding Association Test (WEAT) proposed by \newcite{caliskan2017semantics}. WEAT assesses the extent to which word embeddings represent certain cultural biases. The inspiration for the WEAT analysis is grounded in psychological theory as an extension of the \textit{Implicit Association Test}, used to measure bias in humans \cite{iat}. WEAT calculates the semantic similarity between two sets of target words (e.g., male vs. female names) and two sets of attribute words using word embeddings (e.g., career vs. family). Table \ref{tab:weat} indicates the nine WEAT tests and their corresponding targets and attributes.

\begin{table*}[!htbp]
\centering
\resizebox{0.8\textwidth}{!}{
\begin{tabular}{@{}rrll@{}}

\multicolumn{1}{c}{\textbf{Bias}} & \multicolumn{1}{c}{\textbf{\#}} & \multicolumn{1}{c}{\textbf{Targets}} & \multicolumn{1}{c}{\textbf{Attributes}} \\ \hline
\multirow{3}{*}{Conceptual} & 1 & Flowers vs. Insects   & Pleasant vs. Unpleasant  \\
       & 2 & Instruments vs. Weapons  & Pleasant vs. Unpleasant  \\
       & 9 & Mental vs. Physical Disease & Temporary vs. Permanent  \\ \hline
\multirow{3}{*}{Racial}  & 3 & Native vs. Foreign Names  & Pleasant vs. Unpleasant  \\
       & 4 & Native vs. Foreign Names (v2) & Pleasant vs. Unpleasant  \\
       & 5 & Native vs. Foreign Names (v2) & Pleasant vs. Unpleasant (v2) \\ \hline
\multirow{3}{*}{Gender}  & 6 & Male vs. Female Names   & Career vs. Family   \\
       & 7 & Math vs. Arts     & Male vs. Female Terms  \\
       & 8 & Science vs. Arts    & Male vs. Female Terms  \\ \hline
\end{tabular}}
\caption{Overview of our proposed measured bias categories (conceptual, race, and gender) for the WEAT analysis. WEAT compares the association between two different target word lists (i.e. Math vs. Arts) to attribute word lists (i.e. Male vs. Female terms). \# indicates the original WEAT test number \cite{caliskan2017semantics}.}
\label{tab:weat}
\end{table*}

 \newcite{kurpicz2020cultural} apply the same concept to three other languages (German, Italian, and Spanish) and adapted and evaluated four WEAT tests for German. The multilingual name adaptations were created by experts examining the census data for popular names from each country of origin and creating word lists for \textit{Male vs. Female} names, as well as \textit{Native vs. Foreign} names (to replace the \textit{European-American vs. African-American} test originally proposed in WEAT). \newcite{kurpicz2020cultural} do not present a translation for a tenth test on ageism proposed by \newcite{caliskan2017semantics}, so we omit it from our study to not combine differing methodologies. In this work, we present German translations for all nine WEAT tests\footnote{The WEAT words not found in \newcite{kurpicz2020cultural}'s study were translated from the English WEAT through DeepL and corrected by two native German speakers.}.

We broadly categorize the WEAT tests into the three main dimensions of bias: Racial, Gender, and Conceptual. This is in accordance with the literature on bias in educational data (e.g., \newcite{Baker2021AlgorithmicEducation}). Our categorization helps language model users to have a big picture understanding of how their model performs (i.e. model X is more biased by gender than race) instead of granular statements (i.e. model X finds male names more associated with career than with family). The groupings are detailed further in Table \ref{tab:weat}, with each category consisting of three tests.

To quantitatively compare across WEAT analyses, we use the metric proposed by \newcite{caliskan2017semantics}. \textit{Effect size} is a normalized measure of the distance between the two distributions of associations and targets, calculated as follows:

$$
 \frac{{mean}_{x \in X} s(x, A, B)-{mean}_{y \in Y} s(y, A, B)}{std_{w \in X \cup Y} s(w, A, B)}
$$

where $X$ and $Y$ are two sets of target words of equal size, $A$, $B$ are two sets of attribute words, and $s(w, A, B)$ measures the association of embeddings of the target word $w$ with the attribute words.

\section{Results on Bias Analysis}

We present results across three stages of the bias pipeline (highlighted in Figure \ref{fig:concept}): (1) a WEAT co-occurrence analysis examining bias directly in the peer-review corpora, (2) an embedding space analysis using a GloVe architecture trained on the peer-review data, (3) an analysis of the three most popular German language models, before and after fine-tuning on the peer-review corpora. 

\subsection{Bias in the Peer-Review Corpus}

In the first experiment, we conduct a WEAT co-occurrence analysis as proposed by \newcite{spliethover-wachsmuth-2020-argument,caliskan2017semantics}. Our aim is to measure the bias present in the raw corpus without the confounding factors of model architecture and pre-existing bias in embeddings. Therefore, this test identifies specific occasions in the text where target words are present in close proximity to attribute words. The neighborhood of proximity can be defined as within the same sentence or within the same review, but the likelihood of a review mentioning different topics over several sentences is significant and we do not want to conflate circumstantial correlation with bias. Therefore, we only examine co-occurrence by sentence.

\textit{We do not find significant results across any of the nine WEAT tests, with only six co-occurrences identified in total across 9,165 peer-reviews\footnote{For the interested reader, the results on all WEAT analyses can be found in the Appendix. Moreover, the peer-review corpus and the code for the conducted tests can be found in our Github repo: \url{https://github.com/epfl-ml4ed/bias-at-a-second-glance}.}.}

%\subsubsection{Subsets of Peer Backward Assessment Ratings}

In line with existing research, we found that the peer backward assessment ratings have a large skew towards positive ratings, with over 50\% of the data residing in points 6 and 7 across all feedback questions asked. Student judgements about helpfulness may be dependent on the review sentiment \cite{Patchan.2018}. Due to this positive skew, we select ratings < 6 as a \textit{low} rating denomination and ratings >= 6 as a \textit{high} rating across 4 reviewer axes: helpfulness, critical aspects, constructive suggestions, and overall quality. Table \ref{tab:subsets} indicates the number of entries in each subset.

\begin{table}[]
\centering
\small
\begin{tabular}{@{}|r|cc|@{}}

\multicolumn{1}{|c|}{\textbf{Subsets}} & \multicolumn{1}{c|}{\textbf{High (\textgreater{}= 6)}} & \textbf{Low (\textless{} 6)} \\ 
Constructive Suggestions & \multicolumn{1}{c|}{5656} & 3509 \\
Critical Aspects   & \multicolumn{1}{c|}{5514} & 3651 \\
Helpfulness    & \multicolumn{1}{c|}{5886} & 3279 \\
Quality     & \multicolumn{1}{c|}{5391} & 3774 \\

\end{tabular}
\caption{Distribution of 9165 student peer-reviews in each reviewer rating subset. The scores have been rated by the receiver of the review. High and Low ratings correspond to scores on the Likert 7-point scale.}
\label{tab:subsets}
\end{table}

We conducted a WEAT co-occurrence analysis across different review rating criteria (quality, critical aspects, helpfulness, constructive) and WEAT target-attribute pairings, as inspired by \newcite{spliethover-wachsmuth-2020-argument}. Examining the overall corpus, none of the subsets are able to identify significant bias. However, tests with ratings of high quality do find minimal instances of bias in test nine (mental vs. physical disease) while the other rating criteria find test nine co-occurrences in their low rating groups (critical aspects, constructive suggestions, helpfulness). Due to the very few co-occurrences present (two instances found in 9,165 reviews, each containing around ten sentences), this difference could be attributed to noise.

\begin{figure*}[ht!]
 \centering
 \includegraphics[width=\textwidth, trim=8 8 8 8,clip]{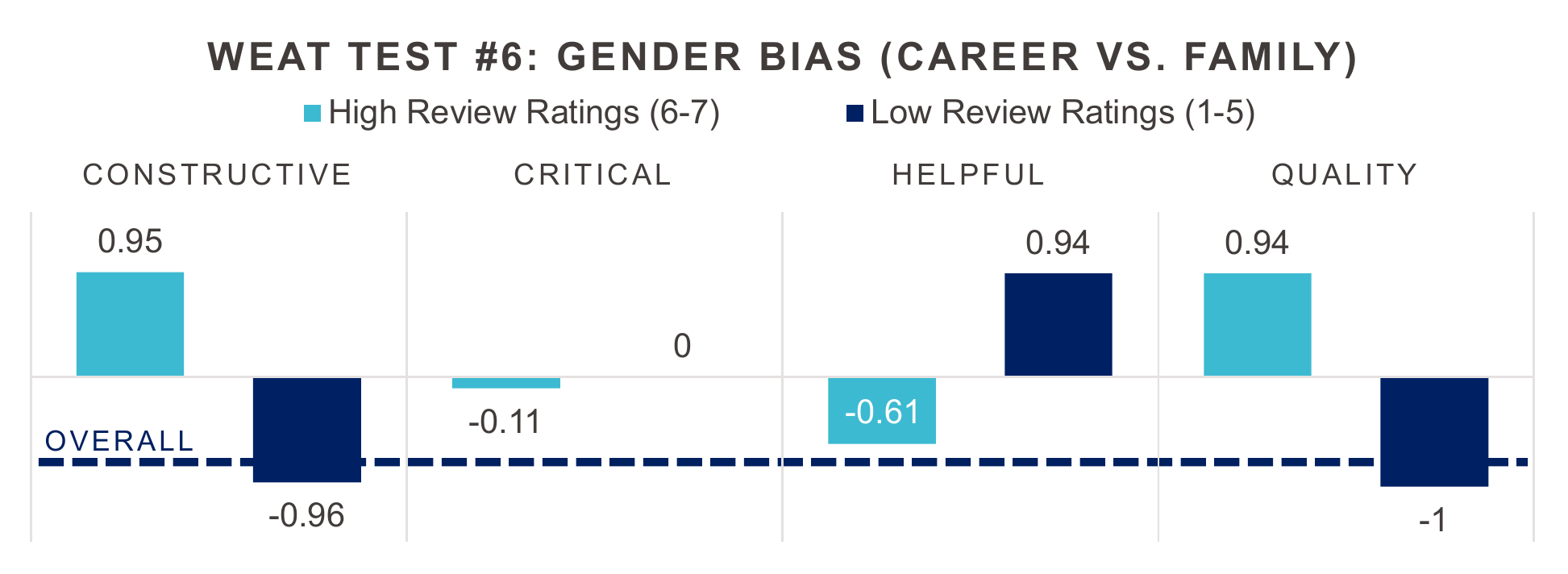}
 \caption{Overview on significant gender bias for WEAT Test 6 in GloVe models between the different review rating subsets. We examine the total effect size as well as each of the subsets (high and low scores for constructive suggestions, critical aspects, helpfulness, and quality). Effect size is normalized between -1.0 to +1.0.}
 \label{fig:glove}
\end{figure*}

% \begin{figure}[ht!]
%  \centering
%  \includegraphics[width=1.05\linewidth, trim=4 4 4 4,clip]{img/gender1.pdf}
%  \caption{ Effect size is normalized between -1.0 to +1.0.}
%  \label{fig:glove-gender}
% \end{figure}

\begin{figure*}[]
 \centering
 \includegraphics[width=\textwidth, trim=4 4 4 4,clip]{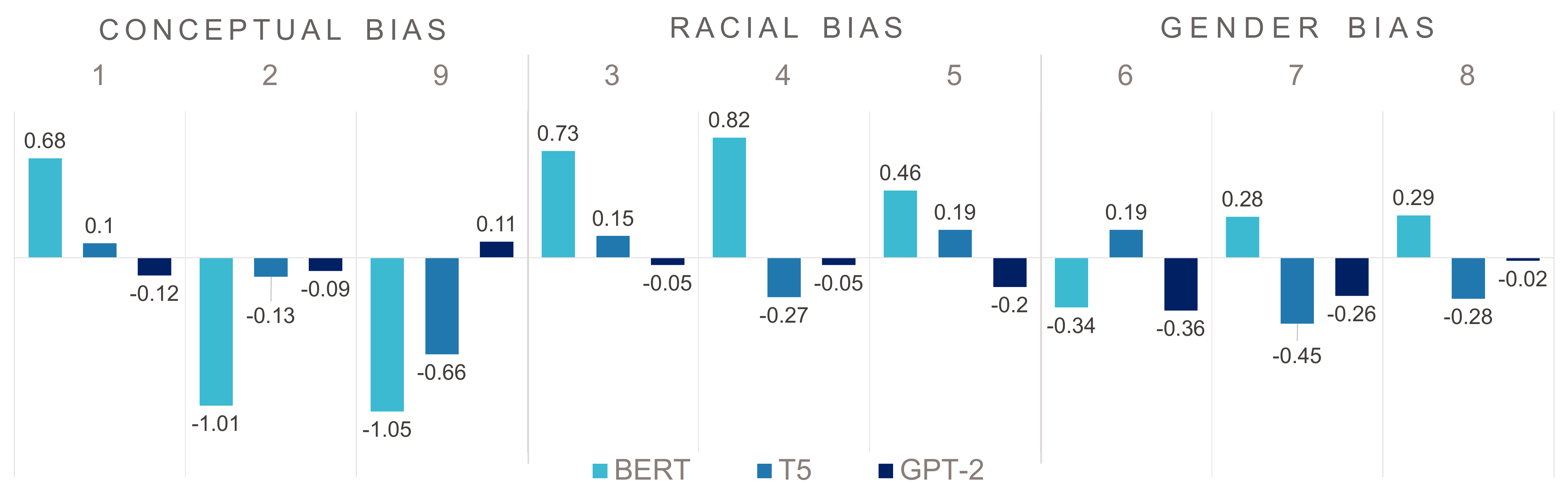}
 \caption{The \textbf{differences in effect size} caused by fine-tuning in comparison with pre-trained model results across all nine WEAT tests on three models (BERT, T5, GPT-2). Effect size is normalized between -1.0 to +1.0.}
 \label{fig:model-results}
\end{figure*}
\textit{In summary, from a preliminary examination of the raw text corpora through the lens of the WEAT co-occurrence tests, we are not able to uncover any significant bias.}
\subsection{Bias in Embeddings}

In a second experiment, we implement a GloVe model trained on the raw text corpora. The data is pre-processed to remove punctuation, stop-words, and HTML tags. We train the GloVe model for 100 epochs, which consisted of about 20 minutes of training time on an 8-core Apple M1 CPU. WEAT Test 6 (Male vs. Female Names :: Career vs. Family) is the only test able to uncover bias. The other eight WEAT tests examined are out-of-vocabulary for the GloVe model. This highlights a distinct disadvantage in training models only on a distinct set of texts instead of leveraging larger language models and adapting them for a certain task. 

Using the entire corpus, we identified a negative bias of -0.748, stating that \textit{female names} are more related to \textit{career} terms than \textit{male names}. Effect sizes are normalized from +1.0 to -1.0, so a bias of 0.75 is very significant. Another equivalent statement is that \textit{male names} are more related to \textit{family} terms than \textit{female names}.

In addition to the overall analysis, we examined subsets of the corpus based on peer backward assessment review ratings classified into high and low rating categories which are shown in Figure \ref{fig:glove}. Eight GloVe models are trained on each subset of the data (i.e. high quality, low helpful). We see that constructive suggestions and quality have the most difference in bias between the ratings with high scores and low scores. In the reviews considered highly constructive and high quality, \textit{male names} are more associated with \textit{career} than \textit{female names}; in their corresponding minimally constructive and low quality counterparts, \textit{female names} are more associated with \textit{career}. This result is notable because it reveals the opposite bias found in these subsets as contrasted with the bias measured over a GloVe model trained on the whole corpus. High and low ratings across critical aspects do not uncover significant differences, but high helpful and low helpful ratings do identify the same bias as the initial GloVe model (\textit{female names} are more associated with \textit{career} than \textit{male names}).

\textit{Overall, only one test on the gender axis is able to uncover bias using traditional word embeddings (GloVe). Comparing bias in a GloVe model trained on the overall corpus and a GloVe model high quality review subsets find different conclusions, emphasizing the importance of granularity in bias analysis. This result is in line with the previous co-occurrence study.}

\subsection{Bias in German Language Models}

In a third experiment, we examined three popular transformer-based German language models for bias: GermanBERT, German T5, and German GPT-2, extracted from the HuggingFace library \cite{wolf2019huggingface}. Few works analyze bias in pre-trained German language models \cite{kraft2021triggering, ahn2021mitigating}, usually referencing one model at a time instead of a comparative study. Therefore, we aim to address this gap in research. Further details on how these models were trained and the fine-tuning objectives can be found in Section \ref{sec:model-arch}.

Our analysis consists of three parts: (1) Conduct WEAT analysis to measure the underlying bias in the pre-trained German models, (2) fine-tune three models on our peer-review text corpora, and (3) measure the change in bias across the WEAT tests. The WEAT scores for pre-trained and fine-tuned models can be found in Appendix Tables \ref{tab:pretrained} and \ref{tab:finetuned}.

We initially conduct the WEAT analysis on the pre-trained language models and find that GermanBERT and German GPT-2 are significantly biased across all three tests on the racial axis (averaging 1.25 and 1.75 in effect size respectively), finding \textit{native names} generally more associated with \textit{pleasant} terms than \textit{foreign} names. German T5 and German GPT-2 are biased across the conceptual axis (averaging 0.38 and 0.51 in effect size respectively), with positive effect sizes for all three tests. 
% \todo{put some case examples of words}

We then pre-process the input data and fine-tune the language models. Figure \ref{fig:model-results} identifies the differences in the WEAT effect sizes across the three axes of bias (nine WEAT tests) after fine-tuning. The gender axis has the least change in score across all three models, showing that fine-tuning on our data does not significantly impact the underlying gender bias in the pre-trained model. However, GermanBERT is highly affected by fine-tuning across the conceptual and racial axes across tests 1-5, and 9. Model T5 is significantly impacted in conceptual test 9 (mental vs. physical disease) and GPT-2 bias results are only minutely impacted by fine-tuning.

We additionally fine-tune the language models with the eight subsets of the ratings, as per the same experiment in Figure \ref{fig:glove}. As we analyze these results, we find the bias does not vary significantly across model subsets. For a point of comparison, we examine WEAT 6 (gender bias across career and family attributes), found as the most significant test in the GloVe model WEAT analysis. We hypothesized that the language model that was least susceptible to fine-tuning (GPT-2) might show stronger variations across subsets, but our results indicate a very small change in bias of at maximum 0.03, with a baseline of 0.61 for the GPT-2 WEAT 6 effect size on the total corpus (Figure \ref{fig:glove}).

\begin{figure*}[ht!]
 \centering
 \includegraphics[width=0.8\textwidth, trim=8 8 8 8,clip]{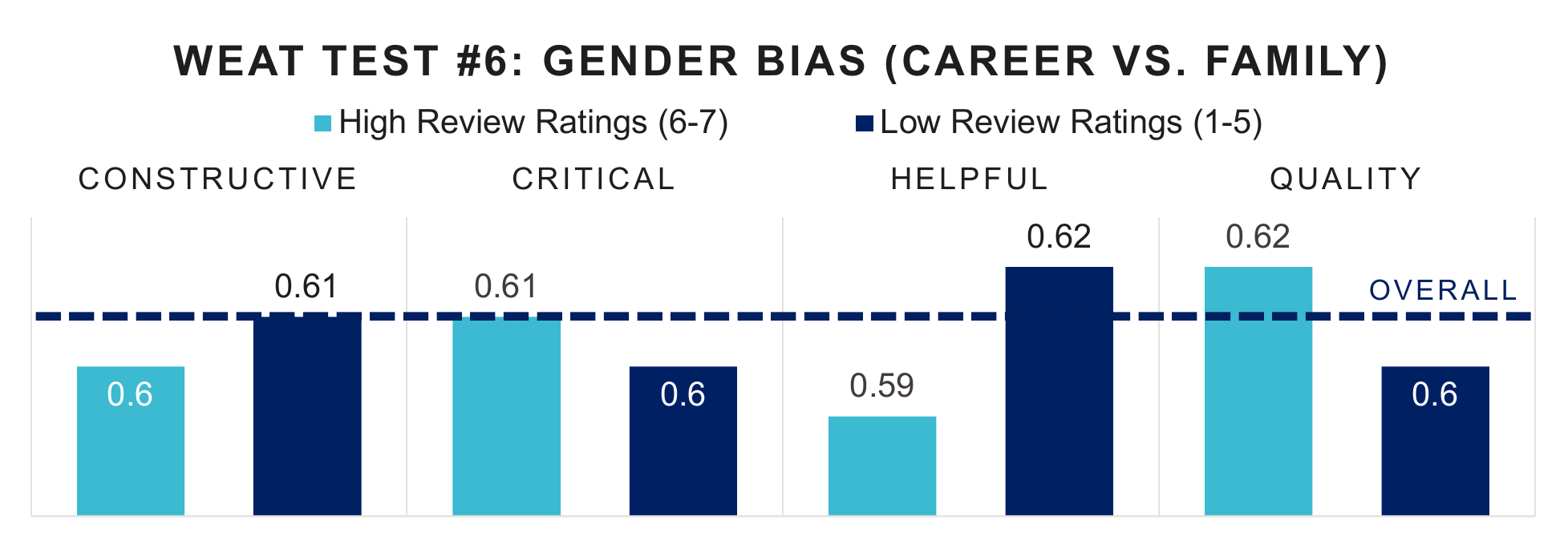}
 \caption{Overview on gender bias (WEAT 6) in \textbf{fine-tuned GPT-2} between the different review rating subsets. Effect size is normalized between -1.0 to +1.0.}
 \label{fig:glove}
\end{figure*}

Moreover, we controlled for different subsets concerning the male or female authors in terms of bias along the NLP pipeline in the last three years of our corpus. Nevertheless, we did not find any significant results in 1) the co-occurrence analysis for gender-separate subsets, 2) for the GloVe embeddings, and 3) for the fine-tuning on BERT, T5 and GPT-2. The exact results for the fine-tuned models can be found in Tables \ref{tab:gender1} and \ref{tab:gender2}.

\begin{table}[!h]
\small
\centering
\resizebox{0.8\linewidth}{!}{
\begin{tabular}{@{}cccc@{}}
 & \textbf{\begin{tabular}[c]{@{}c@{}}Overall\\ (2017-2019)\end{tabular}} & \textbf{\begin{tabular}[c]{@{}c@{}}Male \\ Authors\end{tabular}} & \textbf{\begin{tabular}[c]{@{}c@{}}Female\\ Authors\end{tabular}} \\ 
\multicolumn{1}{|l|}{\textbf{\begin{tabular}[c]{@{}r@{}}GloVe\\ WEAT 6\end{tabular}}} & \multicolumn{1}{r|}{-0.91} & \multicolumn{1}{r|}{0.39} & \multicolumn{1}{r|}{0} \\ 
\end{tabular}}
\caption{WEAT results for the GloVe embeddings across subsets of male and female authors. Only WEAT 6 is found significant.}
\label{tab:gender1}
\end{table}

\begin{table*}[!h]
\centering
\resizebox{0.8\textwidth}{!}{
\begin{tabular}{rrrrrrrrrrr}
\multicolumn{1}{c}{\textbf{}}                                     & \multicolumn{1}{c}{\textbf{}}    & \multicolumn{3}{c}{\textbf{Conceptual}}                  & \multicolumn{3}{c}{\textbf{Racial}}                    & \multicolumn{3}{c}{\textbf{Gender}}                    \\ 
\multicolumn{1}{|c|}{\textbf{Model}}                                         & \multicolumn{1}{l|}{\textbf{Author}} & \multicolumn{1}{c}{1} & \multicolumn{1}{c}{2} & \multicolumn{1}{c|}{9}  & \multicolumn{1}{c}{3} & \multicolumn{1}{c}{4} & \multicolumn{1}{c|}{5}   & \multicolumn{1}{c}{6} & \multicolumn{1}{c}{7} & \multicolumn{1}{c|}{8}   \\ \hline
\multicolumn{1}{|r|}{\multirow{2}{*}{\begin{tabular}[c]{@{}r@{}}German\\ BERT\end{tabular}}} & \multicolumn{1}{r|}{male}      & -0.11         & 0.58         & \multicolumn{1}{r|}{0.17} & 0.47         & 0.47         & \multicolumn{1}{r|}{0.66} & 0.62         & 0.41         & \multicolumn{1}{r|}{-0.23} \\
\multicolumn{1}{|r|}{}                                         & \multicolumn{1}{r|}{female}     & -0.11         & 0.58         & \multicolumn{1}{r|}{0.18} & 0.47         & 0.47         & \multicolumn{1}{r|}{0.67} & 0.62         & 0.41         & \multicolumn{1}{r|}{-0.23} \\ \hline
\multicolumn{1}{|r|}{\multirow{2}{*}{\begin{tabular}[c]{@{}r@{}}German\\ T5\end{tabular}}}  & \multicolumn{1}{r|}{male}      & 0.36         & 0.08         & \multicolumn{1}{r|}{0.22} & 0.53         & 0.31         & \multicolumn{1}{r|}{-0.33} & -0.46         & 0.49         & \multicolumn{1}{r|}{0.03} \\
\multicolumn{1}{|r|}{}                                         & \multicolumn{1}{r|}{female}     & 0.35         & 0.09         & \multicolumn{1}{r|}{0.21} & 0.53         & 0.31         & \multicolumn{1}{r|}{-0.35} & -0.46         & 0.51         & \multicolumn{1}{r|}{0.02} \\ \hline
\multicolumn{1}{|r|}{\multirow{2}{*}{\begin{tabular}[c]{@{}r@{}}German\\ GPT-2\end{tabular}}} & \multicolumn{1}{r|}{male}      & 0.07         & 0.11         & \multicolumn{1}{r|}{0.59} & 0.62         & 0.62         & \multicolumn{1}{r|}{0.63} & 0.61         & 0.01         & \multicolumn{1}{r|}{-0.29} \\
\multicolumn{1}{|r|}{}                                         & \multicolumn{1}{r|}{female}     & 0.07         & 0.11         & \multicolumn{1}{r|}{0.59} & 0.62         & 0.62         & \multicolumn{1}{r|}{0.64} & 0.61         & 0.01         & \multicolumn{1}{r|}{-0.29} \\ 
\end{tabular}}
\caption{WEAT analysis from subsets of male authors and female authors used to fine-tune three German language models. Most values remain the same across both subsets at two degrees of precision.} 
\label{tab:gender2}
\end{table*}

\textit{Despite the previous two experiments not finding pervasive bias in the corpora, pre-trained German language models are inherently significantly biased, and fine-tuning using language models uncovers different, significant bias. BERT is the most susceptible to changes in bias of the three architectures.}

\section{Discussion and Conclusions}

We collected and analyzed a novel corpus of 9,165 German peer-reviews, including the students' gender and peer-reviewed helpfulness ratings, to perform a granular bias analysis along the NLP pipeline. Our aim was to shed light on the popular pedagogical scenario of peer-reviewing, where NLP and ML are extensively used for improving adaptivity. Our results did not show any significant bias across any of the nine WEAT tests for our corpora or the collected ratings. For the German GloVe embeddings, we only found a significant gender bias for test 6 involving male and female names associated with career and family. Importantly, in common pre-trained German Language models (BERT, T5, GPT-2), we found substantial conceptual, racial, and gender bias. We saw that after fine-tuning on our corpora, the language models uncovered other significant bias that were not present before fine-tuning. BERT was most susceptible to bias changes. Hence, we contribute a perspective in how to reveal and investigate bias in educational corpora for educational downstream tasks, as well as initial actionable considerations for educational data scientists intending to use our corpus.

Our results add insights to the literature about gender bias in educational data modelling (e.g., \newcite{Anderson2019AssessingTF}), in embedding spaces (e.g., \newcite{bolukbasi2016man}), and in language modelling (e.g., \newcite{Lu2020GenderBI}). In our research, we built on the findings from these different perspectives along the NLP pipeline and conducted a fine-granular analysis for German peer-reviews by analyizing the texts, the qualitative review scores, demographics (i.e, gender), embeddings, and the most common pre-trained language models before and after fine-tuning. %Hence, our research adds an exemplary methodology for investigating bias in educational corpora (see Figure \ref{fig:concept}) and, therefore, adds to past literature on bias analysis along the NLP pipeline (e.g., \newcite{Hovy2021})

Our results suggest three main directions. First, bias can emerge and change along the NLP-pipeline. Detecting a certain bias in the corpora or in pre-trained language models does not necessitate a connection to bias in fine-tuned models for downstream tasks. Thus, it is necessary to have more in-depth analyses of bias not only along the entire NLP pipeline but also for subsets of data that are significant. Second, it is important to examine the bias in pre-trained models before using them. And third, more investigations on multilingual data bias are necessary.
Therefore, we contribute to literature on bias of educational language data by providing a fine-grained analysis of one particular but increasingly used pedagogical scenario (peer-reviews). We contribute our collected corpus of peer-reviews in German for further analysis and hope to provide researchers and practitioners with a detailed analysis and discussion of bias in NLP for education. Finally, we aim to contribute to the UN sustainability goal four for a high quality education and fair (digital) education for all.

\subsection{Ethical Considerations}
We note that this research was conducted by a mixed team of authors with Western European, Indian, North-American, female and male backgrounds.

\bibliography{references_all_2, custom}

\bibliographystyle{acl_natbib}

%\section{Acknowledgements}
%HSG, SERI

\appendix
\clearpage
\onecolumn
\section{Appendix}

\begin{table*}[ht!]
\small
\centering
\resizebox{0.62\textwidth}{!}{
\begin{tabular}{@{}|r|c|c|ccc|ccc|ccc|@{}}
\multicolumn{1}{l}{} & \multicolumn{1}{l}{}& \multicolumn{1}{l}{} & \multicolumn{3}{c}{\textbf{Conceptual}} & \multicolumn{3}{c}{\textbf{Racial}} & \multicolumn{3}{c}{\textbf{Gender}} \\ 
\textbf{} & \textbf{\#} & \textbf{+/-} & \textbf{1} & \textbf{2} & \textbf{9} & \textbf{3} & \textbf{4} & \textbf{5} & \textbf{6} & \textbf{7} & \textbf{8} \\ 
\textbf{Overall} & 9165 & +/- & & & 1 & 1 & 1 & 1 & 2 & & \\ 
\multirow{2}{*}{\textbf{Quality}} & \multirow{2}{*}{5391} & + & & & & 1 & 1 & & 1 & & \\
& & -  & & & 1 & & & 1 & 1 & & \\ 
\multirow{2}{*}{\textbf{Critical}} & \multirow{2}{*}{5514} & + & & & 1 & 1 & 1 & & 1 & & \\
& & - & & & & & & 1 & 1 & & \\ 
\multirow{2}{*}{\textbf{Helpful}} & \multirow{2}{*}{5886} & + & & & 1 & 1 & 1 & & 1 & & \\
 & & - & & & & & & 1 & 1 & & \\ 
\multirow{2}{*}{\textbf{Constructive}} & \multirow{2}{*}{5656} & + & & & 1 & 1 & 1 & & 1 & & \\
& & - & & & & & & 1 & 1 & & \\ 
\end{tabular}}
\caption{WEAT co-occurrence analysis across different review rating criteria (quality, critical aspects, helpfulness, constructive) and tests (1-9). This table represents the counts of co-occurrence examples present in the high ratings ($+$) and low ratings ($-$) for each criteria.}
\label{tab:weatcounts}
\end{table*}

\begin{table*}[ht!]
\resizebox{\textwidth}{!}{
\begin{tabular}{|r|c|r|r|rrr|}

\multicolumn{1}{|c|}{\textbf{Bias}}        & \textbf{\#} & \multicolumn{1}{c|}{\textbf{Targets}} & \multicolumn{1}{c|}{\textbf{Attributes}} & \textbf{\begin{tabular}[c]{@{}r@{}}German \\ BERT\end{tabular}} & \textbf{\begin{tabular}[c]{@{}r@{}}German \\ T5\end{tabular}} & \textbf{\begin{tabular}[c]{@{}r@{}}German \\ GPT-2\end{tabular}} \\ 
\multicolumn{1}{|c|}{}               & 1      & Flowers vs. Insects          & Pleasant vs. Unpleasant         & {\color[HTML]{000000} -0.22}                  & {\color[HTML]{000000} 0.61}                  & {\color[HTML]{000000} 0.25}                   \\
\multicolumn{1}{|c|}{}               & 2      & Instruments vs. Weapons        & Pleasant vs. Unpleasant         & {\color[HTML]{000000} 0.58}                   & {\color[HTML]{000000} 0.11}                  & {\color[HTML]{000000} 0.15}                   \\
\multicolumn{1}{|c|}{\multirow{-3}{*}{Conceptual}} & 9      & Mental vs. Physical Disease      & Temporary vs. Permanent         & {\color[HTML]{000000} 0.16}                   & {\color[HTML]{000000} 0.5}                  & {\color[HTML]{000000} 0.54}                   \\ 
                          & 3      & Native vs. Foreign Names       & Pleasant vs. Unpleasant         & {\color[HTML]{000000} 0.48}                   & {\color[HTML]{000000} 0.44}                  & {\color[HTML]{000000} 0.64}                   \\
                          & 4      & Native vs. Foreign Names (v2)     & Pleasant vs. Unpleasant         & {\color[HTML]{000000} 0.48}                   & {\color[HTML]{000000} 0.44}                  & {\color[HTML]{000000} 0.64}                   \\ 
\multirow{-3}{*}{Racial}              & 5      & Native vs. Foreign Names (v2)     & Pleasant vs. Unpleasant (v2)       & {\color[HTML]{000000} 0.67}                   & {\color[HTML]{000000} -0.38}                 & {\color[HTML]{000000} 0.74}                   \\ 
                          & 6      & Male vs. Female Names         & Career vs. Family            & {\color[HTML]{000000} 0.61}                   & {\color[HTML]{000000} -0.56}                 & {\color[HTML]{000000} 0.79}                   \\
                          & 7      & Math vs. Arts             & Male vs. Female Terms          & {\color[HTML]{000000} 0.4}                   & {\color[HTML]{000000} 0.73}                  & {\color[HTML]{000000} 0.14}                   \\ 
\multirow{-3}{*}{Gender}              & 8      & Science vs. Arts           & Male vs. Female Terms          & {\color[HTML]{000000} -0.24}                  & {\color[HTML]{000000} 0.22}                  & {\color[HTML]{000000} -0.28}                   \\ 
\end{tabular}}
\caption{WEAT Test effect sizes for \textbf{pretrained} German BERT, T5, and GPT-2. Positive scores indicate that Target 1 (i.e. Mental Disease) is more associated with Attribute 1 (i.e. Temporary) than Target 2 (i.e. Physical Disease). An equivalent statement is that Target 2 (i.e. Physical Disease) is more associated with Attribute 2 (i.e. Permanent) than Target 1 (i.e. Mental Disease). Scores scale between +1.0 and -1.0.}
\label{tab:pretrained}
\end{table*}

\begin{table*}[ht!]
\resizebox{\textwidth}{!}{
\begin{tabular}{|r|c|r|r|rrr|}

\multicolumn{1}{|c|}{\textbf{Bias}}        & \textbf{\#} & \multicolumn{1}{c|}{\textbf{Targets}} & \multicolumn{1}{c|}{\textbf{Attributes}} & \textbf{\begin{tabular}[c]{@{}r@{}}German \\ BERT\end{tabular}} & \textbf{\begin{tabular}[c]{@{}r@{}}German \\ T5\end{tabular}} & \textbf{\begin{tabular}[c]{@{}r@{}}German \\ GPT-2\end{tabular}} \\ 
\multicolumn{1}{|c|}{}               & 1      & Flowers vs. Insects          & Pleasant vs. Unpleasant         & {\color[HTML]{00B050} 0.23}                   & {\color[HTML]{00B050} 0.36}                  & {\color[HTML]{C00000} 0.07}                   \\
\multicolumn{1}{|c|}{}               & 2      & Instruments vs. Weapons        & Pleasant vs. Unpleasant         & {\color[HTML]{C00000} 0.07}                   & {\color[HTML]{C00000} 0.05}                  & {\color[HTML]{C00000} 0.11}                   \\
\multicolumn{1}{|c|}{\multirow{-3}{*}{Conceptual}} & 9      & Mental vs. Physical Disease      & Temporary vs. Permanent         & {\color[HTML]{C00000} -0.37}                  & {\color[HTML]{C00000} 0.17}                  & {\color[HTML]{00B050} 0.6}                    \\ 
                          & 3      & Native vs. Foreign Names       & Pleasant vs. Unpleasant         & {\color[HTML]{00B050} 0.85}                   & {\color[HTML]{00B050} 0.52}                  & {\color[HTML]{C00000} 0.62}                   \\
                          & 4      & Native vs. Foreign Names (v2)     & Pleasant vs. Unpleasant         & {\color[HTML]{00B050} 0.89}                   & {\color[HTML]{C00000} 0.31}                  & {\color[HTML]{C00000} 0.62}                   \\
\multirow{-3}{*}{Racial}              & 5      & Native vs. Foreign Names (v2)     & Pleasant vs. Unpleasant (v2)       & {\color[HTML]{00B050} 0.9}                   & {\color[HTML]{00B050} -0.29}                 & {\color[HTML]{C00000} 0.64}                   \\ 
                          & 6      & Male vs. Female Names         & Career vs. Family            & {\color[HTML]{C00000} 0.44}                   & {\color[HTML]{00B050} -0.46}                 & {\color[HTML]{C00000} 0.61}                   \\
                          & 7      & Math vs. Arts             & Male vs. Female Terms          & {\color[HTML]{00B050} 0.54}                   & {\color[HTML]{C00000} 0.51}                  & {\color[HTML]{C00000} 0.01}                   \\
\multirow{-3}{*}{Gender}              & 8      & Science vs. Arts           & Male vs. Female Terms          & {\color[HTML]{00B050} -0.1}                   & {\color[HTML]{C00000} 0.08}                  & {\color[HTML]{00B050} -0.29}                   \\ 
\end{tabular}}
\caption{WEAT Test effect sizes for \textbf{finetuned} German BERT, T5, and GPT-2, in comparison with pretrained results in Table \ref{tab:pretrained}. Positive scores indicate that Target 1 (i.e. Mental Disease) is more associated with Attribute 1 (i.e. Temporary) than Target 2 (i.e. Physical Disease). Green text indicates a positive effect size change due to finetuning, red text indicates a negative change.}
\label{tab:finetuned}
\end{table*}

\end{document}